\documentclass[conference]{IEEEtran}
\usepackage{graphicx}
\usepackage{cite}
\usepackage{amsmath}
\usepackage{amssymb}
\usepackage{algorithm}
\usepackage{algorithmic}
\usepackage{makecell}
\usepackage{xcolor}
\usepackage{mathtools}
\usepackage{url}

\usepackage{amsfonts}
\usepackage{cuted}
\usepackage{epsfig}
\usepackage{cases}
\usepackage{caption}
\usepackage{bm}
\usepackage{blkarray}
\usepackage{multirow}
\usepackage{longtable}
\usepackage{breakcites}

\DeclareMathOperator*{\argmax}{argmax} 
\DeclarePairedDelimiter\ceil{\lceil}{\rceil}
\DeclarePairedDelimiter\floor{\lfloor}{\rfloor}

\begin{document}

\title{Reinforcement Learning Architectures: SAC, TAC, and ESAC}
\author{
    \IEEEauthorblockN{Ala'eddin Masadeh\IEEEauthorrefmark{1}, Zhengdao Wang\IEEEauthorrefmark{2}, Ahmed E. Kamal\IEEEauthorrefmark{2}} \\
    \IEEEauthorblockA{\IEEEauthorrefmark{1}Al-Balqa Applied University, Al-Salt 19117, Jordan
    \\amasadeh@bau.edu.jo} \\
    \IEEEauthorblockA{\IEEEauthorrefmark{2}Iowa State University (ISU), Ames, IA 50011, USA
    \\\{zhengdao, kamal\}@iastate.edu}
}

\maketitle

\begin{abstract}
The trend is to implement intelligent agents capable of analyzing available information and utilize it efficiently. This work presents a number of reinforcement learning (RL) architectures; one of them is designed for intelligent agents. The proposed architectures are called selector-actor-critic (SAC), tuner-actor-critic (TAC), and estimator-selector-actor-critic (ESAC). These architectures are improved models of a well known architecture in RL called actor-critic (AC). In AC, an actor optimizes the used policy, while a critic estimates a value function and evaluate the optimized policy by the actor. SAC is an architecture equipped with an actor, a critic, and a selector. The selector determines the most promising action at the current state based on the last estimate from the critic. TAC consists of a tuner, a model-learner, an actor, and a critic. After receiving the approximated value of the current state-action pair from the critic and the learned model from the model-learner, the tuner uses the Bellman equation to tune the value of the current state-action pair. ESAC is proposed to implement intelligent agents based on two ideas, which are lookahead and intuition. Lookahead appears in estimating the values of the available actions at the next state, while the intuition appears in maximizing the probability of selecting the most promising action. The newly added elements are an underlying model learner, an estimator, and a selector. The model learner is used to approximate the underlying model. The estimator uses the approximated value function, the learned underlying model, and the Bellman equation to estimate the values of all actions at the next state. The selector is used to determine the most promising action at the next state, which will be used by the actor to optimize the used policy. Finally, the results show the superiority of ESAC compared with the other architectures.

\end{abstract}


\section{Introduction} \label{sec:intro}

In the framework of artificial intelligence (AI), one of the main goals is to implement intelligent
agents with high level of understanding and making correct decisions. These agents should have the capability
to interact with their environments, collect data, process it, and improve their performance with time.

Implementing autonomous agents
capable of learning effectively has been a challenge for a long time \cite{V_Imp_Survey_Deep_RL}.
One of the milestones that have contributed in this field is reinforcement learning (RL).
It is considered as a principled mathematical framework for learning experience-driven
autonomous agents \cite{RL_Book}.
RL has been widely used to implement autonomous agents
\cite{V_Imp_Survey_Deep_RL}, \cite{v_imp_combine_model_based_with_critic},
\cite{v_imp_similar_work_model_actor_critic}, \cite{AlphaGo_zero_main},
\cite{AlphaGo_main}.

RL refers to algorithms enabling autonomous agents to optimize their behavior, and improve
their performance over time. In this context, the agents work in unknown environments, and learns from
trial and error \cite{writing_17}.
RL methods are categorized into two classes, which are model-free learning
and model-based learning. Model-free learning updates the value function after interacting
with an environment without learning the underlying model. On the other hand, model-based
learning estimates the dynamics (the model) of an environment, which is used later to optimize the
policy \cite{sutton_book_2018}, \cite{definition_model_based_model_free}.

Each learning class has its own advantages, and suffers from a number of weaknesses.
Model-based learning is characterized by its efficiency in learning \cite{model_based_strengths_weaknesses_3},
but at the same time, it struggles in complex problems \cite{model_based_strengths_weaknesses_2}.
On the other hand, model-free learning has strong convergence guarantees under certain
situations \cite{RL_Book}, but the value functions change slowly over
time \cite{model_free_limitation}, especially, when the learning rate is small.

The main idea of this work is to merge methods from the two learning classes to implement intelligent agents,
and to overcome the mutual weaknesses of these two classes.
Combining methods from both learning classes has been discussed in many works
\cite{v_imp_combine_model_based_with_critic}, \cite{v_imp_similar_work_model_actor_critic},
\cite{v_imp_model_based_with_model_free}, \cite{v_imp_model_with_actor_critic},
\cite{actor_critic_save_samples_from_transitions}, \cite{v_imp_survey_model_learning},
\cite{dyna_q_dyna_pi}.

In \cite{v_imp_combine_model_based_with_critic}, a method
called model-guided exploration is presented. This method integrates
a learned model with an off-policy learning such as the Q-learning.
The learned model is used to generate good trajectories using trajectory
optimization, and then, these trajectories are mixed with on-policy experience
to learn the agent. The improvement of this method is quite small even when
the learned model is the true model. This returns to the reason of using two
completely different policies for learning.
This also can be explained by the need to learn bad actions too, so that the
agent can distinguish between bad and good actions.
To overcome the weaknesses of the model-guided exploration approach, another method called imagination
rollout was designed \cite{v_imp_combine_model_based_with_critic}.
It was proposed for applications that need large amounts of experience, or when undesirable actions
are expensive. In this approach,
synthetic samples are generated from the learned model that are called the
imagination rollouts. These rollouts, the on-policy samples, and the optimized trajectories
from the learned model are used with
various mixing coefficients to evaluate each experiment. During each experiment, additional
on-policy synthetic rollouts are generated from each visited state
, and the model is refitted.

In \cite{v_imp_model_based_with_model_free}, an algorithm called approximate model-assisted
neural fitted Q-iteration was proposed. Using this algorithm, virtual trajectories are generated
from a learned model to be used for updating the Q function. This work mainly aimes at
reducing the amount of real trajectories required to learn a good policy.

Actor-critic (AC) is a model-free RL method, where the actor learns a policy that is
modeled by a parameterized distribution, while the critic learns a value function and
evaluates the performance of the policy optimized by the actor.
In \cite{v_imp_similar_work_model_actor_critic}, the framework of human-machine nonverbal
communication was discussed. The goal is to enable machines to understand people
intention from their behavior. The idea of integrating AC with model-based learning was proposed.
The learned dynamics of the underlying model are used to control over the temporal difference (TD)
error learned by the critic, and the actor uses the TD error to optimize the policy for exploring
different actions.

In \cite{v_imp_model_with_actor_critic}, two learning algorithms were designed.
The first one is called model learning
actor-critic (MLAC), while the second one is called reference model actor-critic
(RMAC). The MLAC is an algorithm that combines AC with model-based learning.
In this algorithm, the gradient of the approximated state-value function
$\hat{V}(s)$ with respect to the state $s$, and the gradient of the approximated model
$s'=f(s,a)$ with respect to the action $a$ are calculated. The actor is updated by
calculating the gradient of $\hat{V}(s)$ with respect to $a$ using the chain rule and the
previously mentioned two gradients. However, using RMAC, two functions are learned. The first function is the
underlying model. The second one is the reference model $s'=R(s)$, which maps state $s$ to
the desired next state $s'$ with the highest possible value. Then, using the inverse of
the approximated underlying model, the desired action can be found. The integrated paradigm of the
reference model and the approximated underlying model serves as an actor, which maps states
to actions.

One of the promising methods for developing data efficient RL is off-policy RL.
It does not learn from the policy being followed like on-policy methods,
it utilizes and learns from data generated from past interactions with the
environment \cite{off_policy_policy_gradient_01}.
Off-policy learning has been investigated in many works \cite{off_policy_policy_gradient_01},
\cite{Q_learning_Watkins_Dayan}, \cite{off_policy_actor_critic_imp_writing},
\cite{off_policy_Greedy_GQ}.

Q-learning is considered as the most well known off-policy RL method \cite{Q_learning_Watkins_Dayan}.
It enables agents to act optimally in Markovian environments.
This method evaluates an action at a state using its current value, the received
immediate reward resulting from this action, and the value of the expected best action at the next
state. This expected best action at the next state is selected independently of the currently
executed policy, which is the reason for classifying this method as an off-policy learning method.

Combining off-policy learning with AC was studied in \cite{off_policy_actor_critic_imp_writing}.
As mentioned, it is the first AC algorithm that can be applied off-policy, where a target policy
is learned while following and getting data from another behavior policy. Using this algorithm,
a stream of data (a sequence of states, rewards, and actions) is generated according to a fixed
behavior policy. The critic learns off-policy estimates of the value function for the
current actor policy. Then, these estimates are used by the actor for updating the weights of
the actor policy, and so on.

Using off-policy data (generated data from past interaction with the environment)
to estimate the policy gradient accurately was investigated in \cite{off_policy_policy_gradient_01}.
The goal is to increase the learning efficiency, and to reuse past generated data to improve the
performance compared with the on-policy learning.
In \cite{behavior_policy_gradient}, an analytic expression for the optimal behavior policy (off-policy)
is derived. This expression is used to generate trajectories with low variance estimates to improve
the learning process by estimating the direction of the policy gradient efficiently.

In this work, we present a number of proposed RL architectures, which aim at
improving this field, and providing efficient learning architectures that utilize the
available information efficiently. These architectures are called selector-actor-critic (SAC),
tuner-actor-critic (TAC), and estimator-selector-actor-critic (ESAC).
The main contribution is ESAC, which is designed for intelligent agents. It is designed based
on two ideas, the lookahead and intuition \cite{AlphaGo_zero_main},
for environments with Markov decision process (MDP) underlying model. This architecture enables
an agent to collect data from its environment, analyze it, and then optimize its policy to
maximize the probability of selecting the most promising action at each state.
This architecture is implemented by adding two ideas to AC architecture, which are
learning the underlying model and off-policy learning.

This paragraph discusses the main contribution and the differences between our proposed work and
\cite{v_imp_similar_work_model_actor_critic}.  Our architecture, ESAC,
uses the current available
information from the critic and the model learner to estimate the values of all possible actions
at the next state, and then, uses an off-policy policy gradient to optimize its policy before taking an action. 
In contrast, \cite{v_imp_similar_work_model_actor_critic} uses the learned
model and the value received from the critic just to update the current state value.
Using their model, the policy is optimized after selecting an action and experiencing
its return. This may be unwanted, especially, when some actions are bad, expensive,
and their values can be estimated before experiencing them.

The remainder of the paper is organized as follows.
The formulated problem and the actor-critic architecture are presented in Section~\ref{prob_form_and_prev_works}. 
The proposed architectures are described in Section~\ref{The_proposed_models}.
Section~\ref{EASC_discussions} discusses the main differences and properties of the investigated architectures.
Numerical simulation results are presented in Section~\ref{Simulations}.
Finally, the paper is concluded in Section~\ref{Conclusions}.

\section{Actor-critic} \label{prob_form_and_prev_works}
This part reviews the basic AC algorithm for RL. We use
standard notation that is consistent with that used in e.g., \cite{sutton_book_2018}.
Specifically, $s$ and $s'$ denote states, $a$ and $a'$ denote actions,
$Q(s,a)$ denotes the action-value function, and $V(s)$ denotes
the state-value function. The function $\pi(a|s,\boldsymbol{\theta})$ denotes a
stochastic policy function, parameterized by $\boldsymbol{\theta}$.

\begin{figure}[!h]
\centering
\includegraphics[width=2.2 in,height=2.8 in]{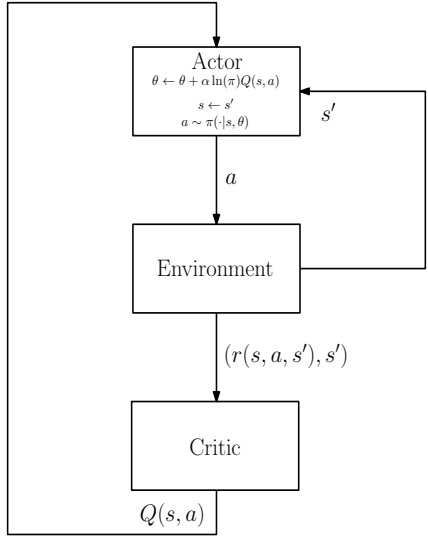}
\caption{Actor-critic architecture.}
\label{actor_critic_fig}
\end{figure}

In AC, the actor generates stochastic actions, and the critic
estimates the value function and evaluate the policy optimized
by the actor. Figure~\ref{actor_critic_fig} shows the interaction between the actor
and the critic in the AC architecture, e.g., \cite{actor_critic_plot}.

In this context, the critic approximates the action-value function $q^{\pi}(s,a)\approx Q(s,a)$, and evaluates
the currently optimized policy using state-action-reward-state-action (SARSA), which is given by
\begin{equation}\label{SARSA_AC}
  Q(s,a) \leftarrow Q(s,a)+\alpha [r(s,a,s')+\gamma Q(s',a')-Q(s,a)],
\end{equation}
where $\alpha$ is the learning rate used to update $Q(s,a)$, $\gamma$ is the discount factor,
and $r(s,a, s')$ is the expected immediate reward resulting from taking action $a$ at state $s$ and transiting
to state $s'$.

The actor uses policy gradient to optimize a parameterized stochastic policy $\pi(a|s,\boldsymbol{\theta})$.
Using policy gradient, the policy objective function $J(\boldsymbol{\theta})$ takes one of three
forms, which are
\begin{itemize}
  \item The value of the start state in episodic environments
  \begin{equation}\label{start_state_J}
    J_1(\boldsymbol{\theta})=V^{\pi}(s_1).
  \end{equation}

  \item The average value in continuing environments
  \begin{equation}\label{avg_J}
    J_{\text{avV}}(\boldsymbol{\theta})= \sum_{s} d^{\pi}(s) V^{\pi}(s).
  \end{equation}

  \item The average reward per time-step in continuing environments also
  \begin{equation}\label{rwd_per_step_J}
    J_{\text{avR}}(\boldsymbol{\theta})= \sum_{s} d^{\pi}(s) \sum_{a} \pi(a|s,\boldsymbol{\theta}) r(s,a),
  \end{equation}
\end{itemize}
where $d^{\pi}(s)$ is the steady-state distribution of the underlying MDP using policy $\pi$, and $r(s,a)$
is the expected immediate reward resulting from taking action $a$ at state $s$. The goal is to
maximize $J(\boldsymbol{\theta})$ \cite{sutton_book_2018}, \cite{gradient_indep_steady_state_1}.
The updating rule for $\boldsymbol{\theta}$ is given by
\begin{equation}\label{updating_rule_theta_J}
  \boldsymbol{\theta} \leftarrow \boldsymbol{\theta}+\beta \nabla_{\boldsymbol{\theta}} J(\boldsymbol{\theta}),
\end{equation}
where $\nabla_{\boldsymbol{\theta}} J(\boldsymbol{\theta})$ is the gradient of $J(\boldsymbol{\theta})$
with respect to $\boldsymbol{\theta}$, and $\beta$ is the step-size used to update the gradient of the policy.

One of the main challenges in this optimization problem is to ensure improvement during changing
$\boldsymbol{\theta}$. This is because changing $\boldsymbol{\theta}$
changes two functions at the same time, which are the policy and the states' distribution.
The other challenge is that the effect of $\boldsymbol{\theta}$ on the states' distribution is unknown,
which makes it difficult to find the gradient of $J(\boldsymbol{\theta})$. Fortunately, policy
gradient theorem provides an expression for the gradient of $J(\boldsymbol{\theta})$ that does not involve the derivative
of the states' distribution with respect to $\boldsymbol{\theta}$ \cite{sutton_book_2018}.
According to policy gradient theorem, for any differentiable policy and for any of the policy objective
function, the policy gradient is \cite{gradient_indep_steady_state_1}
\begin{align}\label{Gradient_J_policy_gradient_thm}
  \nabla_{\boldsymbol{\theta}} J(\boldsymbol{\theta})& \approx  E_{\pi}[ \nabla_{\boldsymbol{\theta}}  \ln(\pi(a|s,\boldsymbol{\theta}))  \, Q(s,a)].
\end{align}

Due to the difficulty of finding the expectation, a stochastic estimate
$\widehat{\nabla_{\boldsymbol{\theta}} J(\boldsymbol{\theta})}$ is used to approximate
$\nabla_{\boldsymbol{\theta}} J(\boldsymbol{\theta})$ \cite{sutton_book_2018}, \cite{Natural_Actor_Critic_Sutton}.
The new updating rule of $\boldsymbol{\theta}$ is given by
\begin{equation}\label{updating_rule_theta_J}
  \boldsymbol{\theta} \leftarrow \boldsymbol{\theta}+\beta \widehat{\nabla_{\boldsymbol{\theta}} J(\boldsymbol{\theta})},
\end{equation}
where
\begin{equation}\label{gradient_estimator_lec_7}
 \widehat{\nabla_{\boldsymbol{\theta}} J(\boldsymbol{\theta})}=\nabla_{\boldsymbol{\theta}} \ln(\pi(a|s,\boldsymbol{\theta})) \, Q(s,a).
\end{equation}

\section{The Proposed Architectures} \label{The_proposed_models}
\subsection{Selector-Actor-Critic} \label{SAC_architecture_subsection}

On-policy learning is defined as methods used to evaluate or improve the same policy used to make
decisions. On the other hand, off-policy approaches try to improve or evaluate a policy different
from the one that is used to generate data \cite{sutton_book_2018}.
This section presents a proposed off-policy policy-gradient method, where the policy being followed is optimized
using the most promising action at state $s$. The idea is to approximate the most promising
action (i.e., the optimal action) at state $s$ by the greedy action $a_g$.
To the best of our knowledge, it is the first work using the most promising action $a_g$ to
optimize stochastic parameterized policies using policy gradient methods. The goal is to optimize the
policy in the direction that maximizes the probability of selecting $a_g$, and increase the speed of
learning a suboptimal $\boldsymbol{\theta}$.
\begin{figure}[!h]
\centering
\includegraphics[width=2.3 in,height=3.3 in]{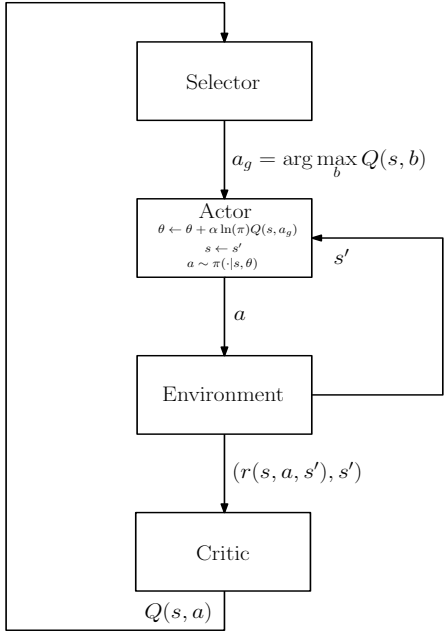}
\caption{Selector-actor-critic architecture.}
\label{off_policy_ctor_critic}
\end{figure}

To achieve this goal, a selector is added to the
conventional AC. Figure~\ref{off_policy_ctor_critic} depicts the SAC model, and the interaction between
its components. The selector determines $a_g$ at the current state greedily according to
\begin{equation}\label{greedy_action}
  a_g= \argmax\limits_{b}  Q(s,b) , ~~ \forall b ~ \text{at} ~ s,
\end{equation}
where $b$ indicates each possible action at state $s$.
After determining $a_g$ by the selector, it is used by the actor to optimize the policy.
The action $a$ selected by the policy being followed in \eqref{gradient_estimator_lec_7}
is replaced by $a_g$.
The new updating rule of $\boldsymbol{\theta}$ is given by
\begin{equation}\label{policy_gradient_theorm_lec_7}
  \boldsymbol{\theta} \leftarrow \boldsymbol{\theta} +\beta \, \nabla_{\boldsymbol{\theta}} \ln(\pi(a_g|s,\boldsymbol{\theta})) \, [Q(s,a_g)].
\end{equation}

After selecting an action using the optimized policy and interacting with the environment, the critic updates the
action-value function according to
\begin{equation}\label{SARSA_SAC}
  Q(s,a) \leftarrow Q(s,a)+\alpha [r(s,a,s')+\gamma Q(s',a')-Q(s,a)].
\end{equation}

\subsection{Tuner-Actor-Critic}\label{TAC_architecture_subsection}
Approximating the underlaying model, and using it with AC learning was discussed
in \cite{v_imp_similar_work_model_actor_critic}. The main idea in \cite{v_imp_similar_work_model_actor_critic}
is to use the learned model to control
over the TD learning, and use TD error to update the policy for exploring different actions. This
section presents our modified architecture, which is called tuner-actor-critic (TAC).
TAC mainly aims at improving the learning process through integrating a tuner and a
model-learner with AC. The main differences between TAC and the proposed model in
\cite{v_imp_similar_work_model_actor_critic} are concluded as follows.
\begin{itemize}
  \item In \cite{v_imp_similar_work_model_actor_critic}, the critic approximates the state-value function
  to evaluate the system performance, while the critic in TAC approximates the action-value function.

  \item In \cite{v_imp_similar_work_model_actor_critic}, the policy uses a preference function for
  selecting actions, which indicates the preference of taking an action at a state. The preference
  function of the current state-action pair is updated by adding its old value to the current TD
  error learned by the critic. On the other hand, the actor in TAC uses stochastic parameterized policies
  to select actions, and uses policy gradient to optimize these policies.

  \item TAC uses the approximated underlying model, the approximated action-value function learned
  by the critic, and the Bellman equation to tune the value of the current state-action pair.
  In contrast, \cite{v_imp_similar_work_model_actor_critic} uses the approximated underlying model
  to find the expected TD error for the current state. The value of the current state is updated
  by adding its previous value to the expected TD error.
\end{itemize}

\begin{figure}[!h]
\centering
\includegraphics[width=3.0 in,height=3.5 in]{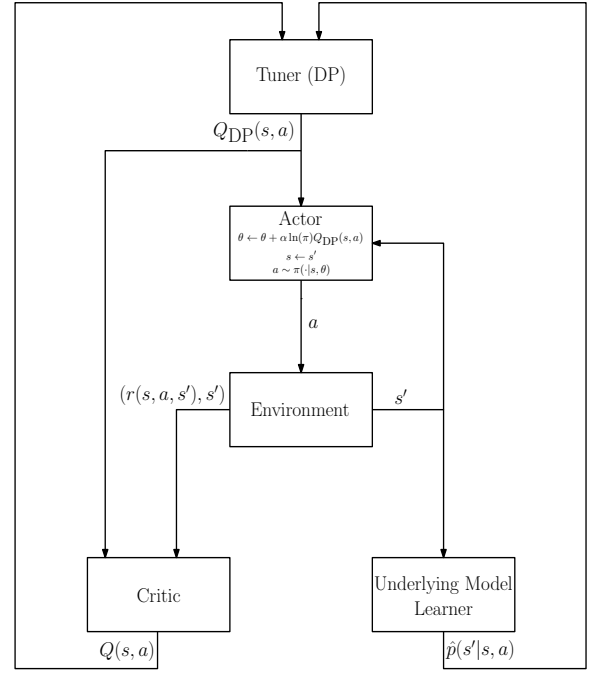}
\caption{Tuner-actor-critic architecture.}
\label{Tuner-Actor-Critic}
\end{figure}

Figure~\ref{Tuner-Actor-Critic} shows the TAC model, and the interaction between
its components. The newly added components to AC architecture are the tuner
and the model learner. Starting from the values received from the critic and the model learner, the
tuner uses the Bellman equation to tune the value of the state-action pair received from the critic.
The tuner tunes the value of $(s,a)$ state-action pair according to
\begin{equation}\label{DP}
  Q_{\text{DP}}(s,a)=\sum_{s'}\hat{p}(s'|s,a)\{r(s,a,s')+ \gamma V(s')\},
\end{equation}
where $\hat{p}(s'|s,a)$ is the approximated probability for transiting from the current state $s$ to next
possible state $s'$ given action $a$ is taken, and $V(s')=\sum_{a'} \pi(a'|s',\boldsymbol{\theta}) Q(s',a')$
is the approximated value of $s'$.

The critic replaces the value of the current state-action pair, $(s,a)$, by the value computed by the
tuner
\begin{equation}\label{SARSA_LHD}
  Q(s,a)\leftarrow Q_{\text{DP}}(s,a).
\end{equation}

The actor updates $\boldsymbol{\theta}$ using
\begin{equation}\label{policy_gradient_theorm_lec_7}
  \boldsymbol{\theta} \leftarrow \boldsymbol{\theta} +\beta \, \nabla_{\boldsymbol{\theta}} \ln(\pi(a|s,\boldsymbol{\theta})) \, [Q_{\text{DP}}(s,a)].
\end{equation}

After selecting an action and interacting with the environment, the critic evaluates the
current policy using
\begin{equation}\label{SARSA_LHD}
  Q(s,a) \leftarrow Q(s,a)+\alpha [r(s,a,s')+\gamma Q(s',a')-Q(s,a)].
\end{equation}

\subsection{Estimator-Selector-Actor-Critic} \label{ESAC_architecture_subsection}
\begin{figure}[h!]
\centering
\includegraphics[width=3.1 in,height=3.5 in]{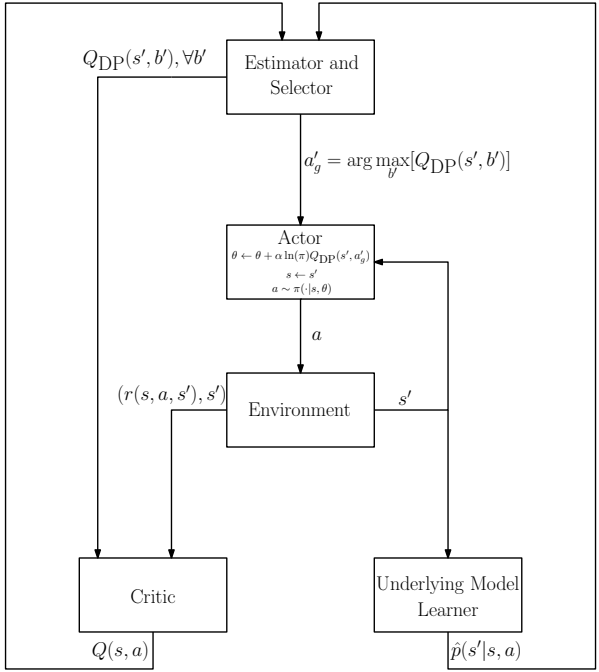}
\caption{Estimator-selector-actor-critic architecture.}
\label{model_actor_critic}
\end{figure}

This architecture aims at providing an intelligent agent. It enables agents to lookahead
in unknown environments by estimating the values of the available actions at the next
state, before optimizing the policy and taking an action. It optimizes the policy based
on the estimated values of the actions at the next state instead of the value of the
experienced action at the current state. This enables agents to maximize the probability
of selecting the most promising action at the next state before taking an action. 
This is the main contribution in this paper, and the main property that distinguishes ESAC
from AC, TAC, and SAC, which optimize the policy based on the experienced action at the current state.
ESAC mainly consists of a model learner, estimator and selector,
an actor, and a critic. Figure~\ref{model_actor_critic} shows the interaction between these components.

The tuner in TAC is renamed as estimator in ESAC. The reason for renaming this part is explained as
follows. In TAC, this part is just used to tune the value of the current state-action pair approximated
by the critic. However, ESAC uses this part to estimate the values of all actions at the next state.
This step shows the look-ahead capability of this model.

Using Bellman equation, and the last updates from the critic and the model learner,
the estimator estimates the values of all the available actions at the next state $s'$
according to
\begin{equation}\label{LHD}
  Q_{\text{DP}}(s',b')=\sum_{s''}\hat{p}(s''|s',b')\{r(s',b',s'')+ \gamma V(s'')\}, ~~ \forall b' ~~ \text{at} ~~ s',
\end{equation}
where $s''$ refers to next possibly reachable
states from state $s'$ given action $b'$, and $\hat{p}(s''|s',b')$ is the
approximated probability for transiting from $s'$ to $s''$
given action $b'$ is taken. Then, the selector determines
the most promising action $a'_g$ at $s'$ to be used by the actor to optimize the policy.
The most promising action at $s'$ is given by
\begin{equation}\label{Q_max}
  a'_{g}=\argmax_{b'} [Q_{\text{DP}}(s',b')], ~~~~~ \forall b' ~~ \text{at} ~~ s'.
\end{equation}

The critic updates the values of actions at $s'$ according to the last update from the
estimator using
\begin{equation}\label{SARSA_LHD}
  Q(s',b')\leftarrow Q_{\text{DP}}(s',b'), ~~~~~~~~ \forall b' ~~ \text{at} ~~ s'.
\end{equation}

The actor updates $\boldsymbol{\theta}$ according to
\begin{equation}\label{policy_gradient_theorm_lec_7}
  \boldsymbol{\theta} \leftarrow \boldsymbol{\theta} +\beta \, \nabla_{\boldsymbol{\theta}} \ln(\pi(a'_g|s',\boldsymbol{\theta})) \, [Q_{\text{DP}}(s',a'_g)].
\end{equation}
where this step shows the intuition capability provided by ESAC, which is to
use the most promising action at next state $s'$ to optimize the policy.

After selecting an action and interacting with the environment, the critic updates the
action-value function according to
\begin{equation}\label{SARSA_LHD}
  Q(s,a) \leftarrow Q(s,a)+\alpha [r(s,a,s')+\gamma Q(s',a')-Q(s,a)].
\end{equation}

\section{Discussions} \label{EASC_discussions}
This paper discusses a number of RL architectures. The first
architecture is called actor-critic (AC). It mainly consists of an actor and a critic.
The actor uses a stochastic parameterized policy to select actions, and policy gradient
to optimize the policy.
The critic approximates a value function, and evaluates the optimized policy by the actor.

The second architecture is called selector-actor-critic (SAC). The newly added component is the
selector. In AC architecture, the actor uses the action selected by the current policy at the current
state to optimize the policy's parameters. However, the selector in SAC determines the most promising
action at the current state, which is used by the actor to optimize the policy's parameters.

The third scheme is called tuner-actor-critic (TAC). It has two more elements added to AC, which
are a model learner and a tuner. The model learner approximates the dynamics of the underlying
environment, while the tuner tunes the value of the current state-action pair using the Bellman
equation, the learned model, and the learned value function by the critic. The actor uses the
tuned value of the current state-action pair to optimize the policy's parameters.

The last model is called estimator-selector-actor-critic (ESAC). The new components added to AC
are a model learner, an estimator, and a selector. Before selecting an action, the estimator
estimates the values of available actions at the next state using the Bellman equation, the learned
model, and the learned value function. Then, the selector determines the most promising action at
the next state, which is used by the actor to optimize the policy. This model mimics rational humans
in the way of analyzing the available information about the next state before taking an action. It aims
at maximizing the probability of selecting the most promising action, and minimizing the probability of
selecting bad and dangerous actions at the next state.

\section{Experimental Results}\label{Simulations}
This section evaluates the proposed architectures.
To evaluate these architectures, we use Value iteration (VI) to find the optimal solution as a benchmark
when possible using the true underlying model \cite{MDP_value_iteration_access_point_selection}.
AC algorithms from \cite{sutton_book_2018}, \cite{v_imp_similar_work_model_actor_critic} are also
compared with.

\subsection{Experimental Set-up}

Two scenarios were considered in the simulation; a simple scenario with small number of states,
and a scenario with large number of states.
The simple scenario is used to evaluate and compare the proposed architectures with the optimal performance
and AC.
For the large scenario, the proposed models are only compared
with AC, where it is difficult to find the optimal solution.

In all scenarios, the discount factor $\gamma$ is set to $0.9$.
The learning rate $\alpha$ used by the critic is set to $0.1$.
The step-size learning parameter $\beta$ used in
policy gradient is set to $0.1$. All the simulations started with an initial policy selecting the available
actions uniformly. The approximated transition model was initialized
with zero transition probabilities.

To evaluate the performance of the considered architectures, a number of MDP problems with
different number of states and different dynamics were considered. The goal is to maximize the
discounted return,
where the discounted return following time $t$, $G_t$, is given by
\begin{equation}\label{cumulative_reward}
  G_t=\sum_{i=t}^{T-1}\gamma^{i-t} R_{i+1},
\end{equation}
where $t$ is the starting time for collecting a sequence of rewards,
$T$ is a final time step of an episode.

In the simulated environments, the simple scenario is modeled by an MDP
with 18 states. Three actions are available with different immediate
rewards and random transition probabilities.
The second scenario is modeled using 354 states. The available actions are 7 with different
immediate rewards and random transition probabilities. All the results were averaged over 500 runs.
The starting state is selected randomly,
where all the states have equal probability to be
the starting state. All mentioned parameters
were used in all experiments unless otherwise stated. More details about the parameters used
in the simulation are available in Appendix A.

\subsection{Exponential Softmax Distribution}
In this work, the exponential softmax distribution \cite{sutton_book_2018} is used as
a stochastic policy to select actions at states. The policy is given by
\begin{equation}\label{softmax}
  \pi(a|s,\boldsymbol{\theta}_a^s)=\frac{\exp(h(s,a,\boldsymbol{\theta}_a^s))}{\sum_{b}\exp(h(s,b,\boldsymbol{\theta}_b^s))},
\end{equation}
where $\exp(x)$ is the base of the natural logarithm, $h(s,a,\boldsymbol{\theta}_a^s)\in \mathbb{R}$
is the parameterized preference for $(s,a)$ pair, and $\boldsymbol{\theta}_a^s$ is the policy's parameter related
to action $a$ at state $s$. For discrete and small action spaces, the
parameterized preferences can be allocated for each state-action pair \cite{sutton_book_2018}.

The parameterized preferences are functions of feature functions $\boldsymbol{\phi}(s,a)$ and the vector
$\boldsymbol{\theta}_a^s$, which are used to determine the preference of each action at each state.
The action with the highest preference at a state will be selected with the highest
probability, and so on \cite{sutton_book_2018}. These preferences can take different forms.
One of the simple forms is that when the preference is a linear function of the weighted
features, which is given by
\begin{equation}\label{linear_preference}
  h(s,a,\boldsymbol{\theta}_a^s)=\boldsymbol{\theta}_a^{s\top} \boldsymbol{\phi}(s,a).
\end{equation}

The $\nabla_{\boldsymbol{\theta}_a^s} \ln(\pi(a|s,\boldsymbol{\theta}_a^s))$ is given by
\begin{align}\label{softmax_update}
  \nabla_{\boldsymbol{\theta}_a^s} \ln(\pi(a|s,\boldsymbol{\theta}_a^s))&=\frac{\nabla \pi(a|s,\boldsymbol{\theta}_a^s)}{\pi(a|s,\boldsymbol{\theta}_a^s)} \\ \nonumber
  & =\boldsymbol{\phi}(s,a)-\pi(a|s,\boldsymbol{\theta}_a^s) \, \boldsymbol{\phi}(s,a).
\end{align}

The feature function $\boldsymbol{\phi}(s,a)$ for $(s,a)$ pair is used for representing
the states and actions in an environment. Feature functions should correspond to aspects
of the state and action spaces, where the generalization can be implemented properly
\cite{sutton_book_2018}.
This work uses binary feature functions. Feature function for a state-action pair is
set to one if action $a$ satisfies the feasibility condition at state $s$, otherwise,
it is set to zero.

\subsection{Comparisons} \label{normal_comparison_small}

In this experiment, the discounted return $G_t$ of each architecture was evaluated.
The optimal performance uses the optimal policy from the first time slot. It requires
a priori statistical knowledge about the environment, which is unavailable to the
remaining architectures. Value iteration (VI) was used to find the optimal policy to find the
upper-bound \cite{MDP_value_iteration_access_point_selection}.

\begin{figure}[H]
\centering
\includegraphics[width=3.4in,height=2.8in]{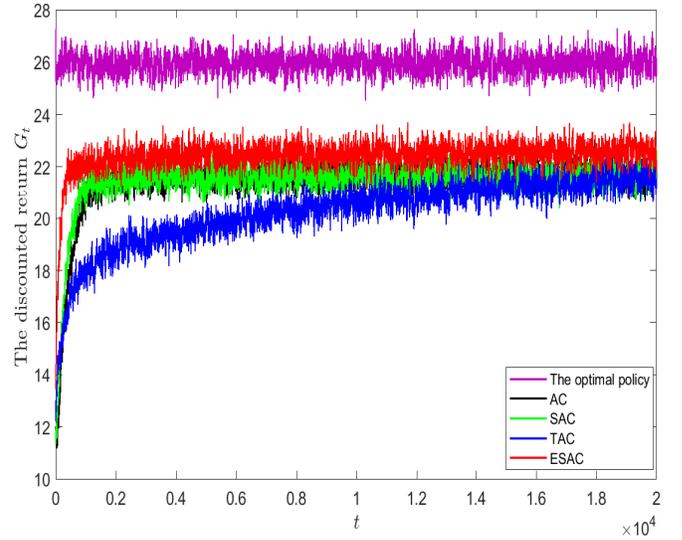}
\caption{The discounted return $G_t$ versus $t$.}
\label{G_t_gamma_0_9_small}
\end{figure}

Figure~\ref{G_t_gamma_0_9_small} shows the discounted return $G_t$ versus $t$ for the considered
architectures. As expected, the discounted return of the optimal policy takes a near-constant pattern from
the first time slot. This is due to using one policy all the time, and the discount factor $\gamma$ which restricts
the discounted return to a certain value. The discounted returns of the RL architectures increase with
experience significantly, in the beginning. As the time increases, they start taking a
near-constant pattern, which results from learning policies that could not be improved
any more, and $\gamma$ that restricts the discounted return of the architectures to certain values.
As shown, ESAC has found a suboptimal policy before AC, TAC, and SAC. Explanations for these results are
summarized as follows. AC, SAC, and TAC are risky architectures, and they do not have estimations
about actions in the beginning.
They need to experience different actions to get accurate estimations about their values to optimize
their policies. This means experiencing different actions, in the beginning, including low-value actions that result in relatively
low discounted returns.
AC experiences an action at the current state, and then, it optimizes the policy based on the approximated
value of the current state-action pair. TAC just tunes the approximated value of the experienced action using
the learned underlying model, then, this tuned value is used by the actor to optimize the policy. It is
clear that both AC and TAC do not exploit the available information about the remaining actions at the current
state to optimize the policy. SAC experiences an action at the current state, and then, based on its approximated
value and the approximated values of other actions, it optimizes the policy.
The superiority of ESAC in finding a better suboptimal policy in a shorter time compared with
the remaining approaches without taking a risky path is
due to its capability to utilize information from other states, and use this information to estimate the most promising
action at next state before optimizing the policy and experiencing an action.

\begin{figure}[H]
\centering
\includegraphics[width=3.4in,height=2.85in]{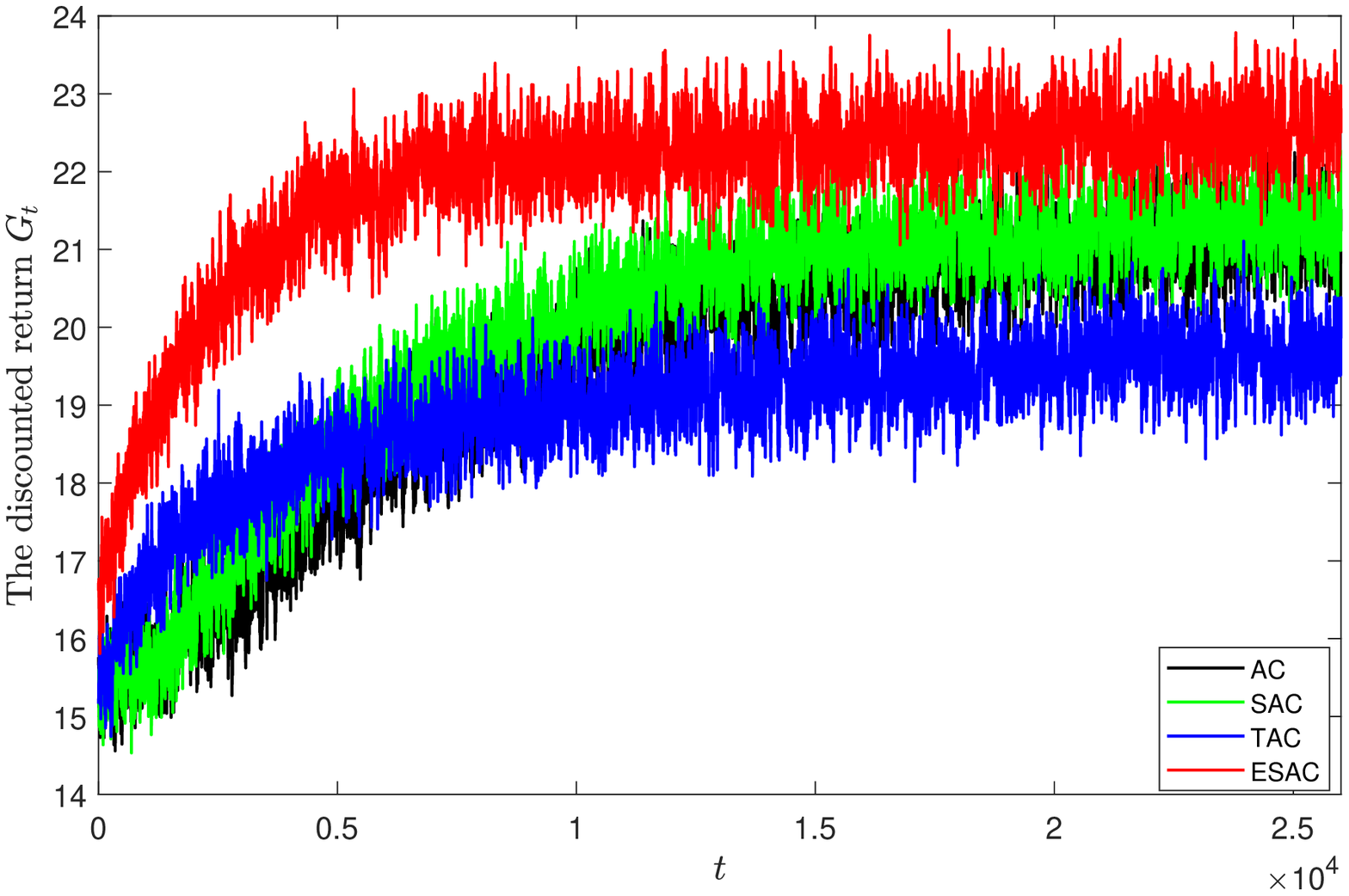}
\caption{The discounted return $G_t$ versus $t$.}
\label{comparison02}
\end{figure}

Figure~\ref{comparison02} shows the performances
of ESAC, SAC, TAC, and AC when there are 364 states.
It shows the superiority of ESAC compared to other competitors even
in the case of large number of states.
It can also be observed that TAC has a faster
initial learning rate, and SAC converges to a higher return,
both compared to AC.

\subsection{Rarely visited states}\label{comp_rar}
We investigate the performance of different models in the
case of having rarely visited good states. For such cases, the
opportunity to increase the cumulative reward is small. Also,
experiencing bad actions would be very expensive. Optimizing
the policy and selecting an optimal action at rarely visited
states is difficult due to lack of experience at these states.
So, the available information should be utilized efficiently to
make correct decisions. This leaves room for improving the
performance based on previous experience.

ESAC utilizes information from other states and the approximated underlying model to estimate
the value of an action. This enables agents to estimate the actions' values at the next state
even if it is visited rarely or if it has not been visited before, optimize the policy before
taking an action, and select appropriate actions at rare good states. However, AC, TAC, and SAC
experience an action, then, the actor optimizes the policy based on the action's return. This may
prevent exploiting rare good states efficiently, especially, when the actions' values can be
estimated accurately before optimizing the policy and selecting an action.

\begin{figure}[h!]
\centering
\includegraphics[width=3.4in,height=2.7in]{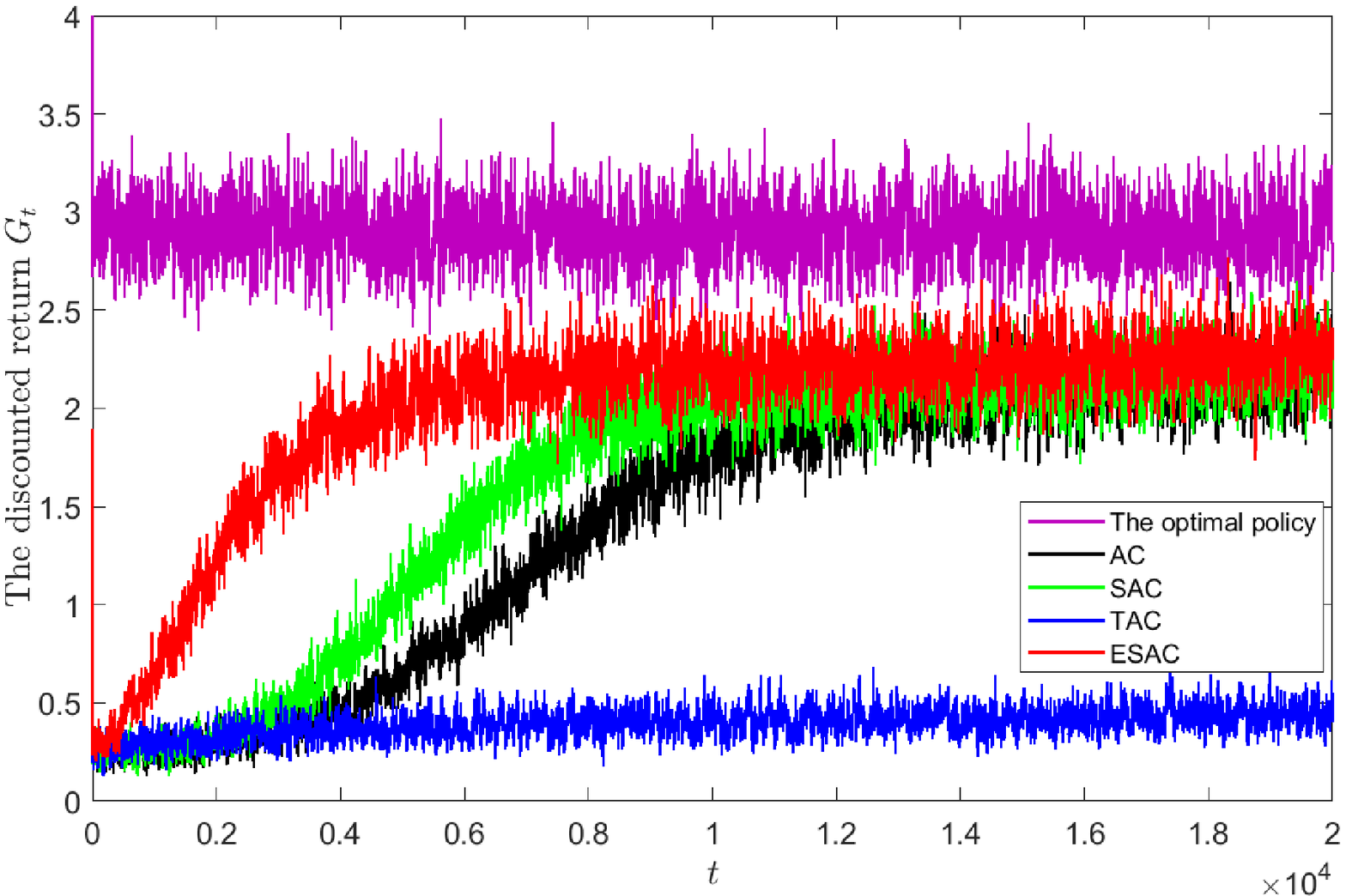}
\caption{The discounted return $G_t$ versus $t$.}
\label{comparison03}
\end{figure}

\begin{figure}[h!]
\centering
\includegraphics[width=3.4in,height=2.7in]{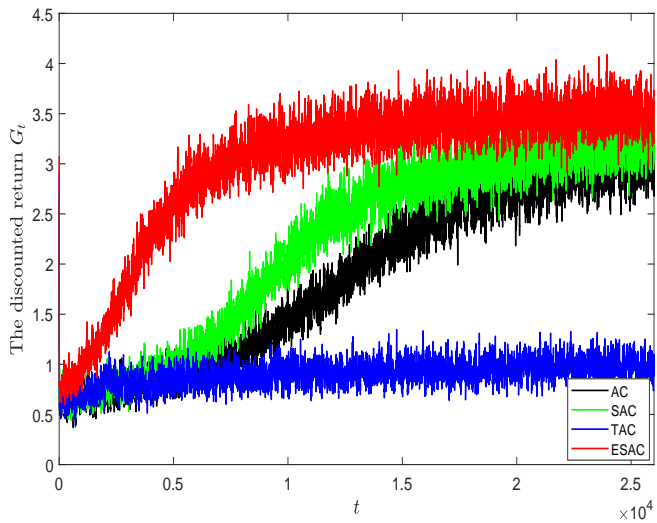}
\caption{The discounted return $G_t$ versus $t$.}
\label{comparison04}
\end{figure}

Figure~\ref{comparison03} and Figure~\ref{comparison04} show the performance
of different architectures for scenarios with 18 and 364 states, respectively, when good
states are visited rarely. Regarding to the quality and the speed of finding
a suboptimal policy, the results show the superiority
of ESAC in environments with rarely visited good states.
The results also show that the SAC outperforms TAC and regular AC.

\section{CONCLUSIONS}\label{Conclusions}

In this paper, three new RL algorithms, named selector-actor-critic,
tuner-actor-critic and estimator-selector-actor-critic are proposed
by adding components to the conventional AC algorithm. Instead of using on-policy
policy gradient as in AC, SAC uses off-policy policy
gradient. TAC aims at improving the learned value function
by adding a model-learner and a tuner to improve the learning
process. The tuner tunes the approximated value of the current
state-action pair using the learned underlying model and the
Bellman equation. AC, SAC, and TAC experience an action,
and then, they optimize their policies based on the value of
the experienced action.
The goal of developing ESAC is to provide an RL architecture for intelligent agents that mimic human
in the way of making decisions. It aims at maximizing the probability of selecting promising actions,
and minimizing the probability of selecting dangerous and expensive actions before selecting actions.
It takes a safe path in optimizing its policy. The ESAC architecture uses two ideas, which are the lookahead
and intuition, to implement such agents with high level of understanding and analyzing available information
before making decisions. The lookahead appears in collecting information from the model learner and the critic
to estimate the values of all available actions at next state. The intuition is seen in optimizing
the policy to maximize the probability of selecting the most promising action. To maintain exploration
during learning, ESAC does not select the most promising action each iteration, it just maximizes its
probability of being selected.
Simulation results also show that
ESAC outperforms all other potential competitors in terms of the discounted return.
This is due to the conservative behavior of ESAC, which prefers to take a
safer path from the beginning compared to the remaining architectures. AC, SAC, and TAC experience
an action, and then, they optimize their policies based on the value of the experienced action.
This dangerous behavior might be unwanted in some
applications, where experiencing dangerous actions to evaluate them is expensive such as learning robots
and drones. It can also be observed that SAC
seems to offer higher converged returns, and TAC is preferred
if faster initial learning rate is desirable, both
compared to AC.

\bibliographystyle{IEEEtran}
\bibliography{refs_model_actor_critic_ICASSP}

\onecolumn{ \section{Appendix A}
 \subsection{Small Scenario - 18 States}
The small scenario with 18 states was implemented as follows. The set of the states is
$S=\{0,...,17\}$. The states evolves according to:

{
\small
\begin{numcases}{s'=}
   6\floor*{\frac{s}{6}}-6a+w, & $s$  is even or 0  \nonumber \\
   6\floor*{\frac{s}{6}}+w, & $s$  is odd and greater than 12 and  $a=0$ \nonumber \\
   6\ceil*{\frac{s}{6}}-6a+w, & else  \label{negative}
\end{numcases}
}

The disturbance is modeled as a Markov process with values in
$\mathcal{W}=\{0,...,5\}$, and transition probability matrix $P$.

The transition probability from current state $s$ to next state $s'$, given that action $a$
is selected, is given by
{
\small
\begin{equation} \label{state_transition_prob}
p(s'|s,a)= \begin{cases}
      p(w|w^p), & \text{if}~ \eqref{negative}~ \text{is satisfied} \\
      0, & \text{else}
   \end{cases}
\end{equation}
}
where $w$ and $w^p$ are the current and the previous values of the disturbance, respectively.

The set of actions at state $s$ is given by $\mathcal{A}^s=\{0,..,a^s_{\max}\}$,
where $a^s_{\max}=\floor*{\frac{s}{6}}$.

In this experiment,
the immediate rewards are determined by the current state and the taken action regardless
the next state. The reward matrix is given by
\[R=
\scriptsize
\begin{blockarray}{cccc}
        & a=0 & a=1 & a=2  \\
\begin{block}{c(ccc)}
         s=0   &  0.0000   &         -                     &        -    \\
         s=1   &  0.0000   &         -                     &        -    \\
         s=2   &  0.0000   &         -                     &        -   \\
         s=3   &  0.0000   &         -                     &        -   \\
         s=4   &  0.0000   &         -                     &        -   \\
         s=5   &  0.0000   &         -                     &        -   \\
         s=6   &  0.0000   &       0.0000                  &        -   \\
         s=7   &  0.0000   &       0.0000                  &        -   \\
         s=8   &  0.0000   &     1.6326 \times 10^{(-07)}  &        -   \\
         s=9   &  0.0000   &     1.6326 \times 10^{(-07)}  &        -   \\
         s=10   &  0.0000   &       8.5067                  &        -   \\
         s=11   &  0.0000   &       8.5067                  &        -   \\
         s=12   &  0.0000   &       0.0000                  &     0.0000 \\
         s=13   &  0.0000   &       0.0000                  &     0.0000 \\
         s=14   &  0.0000   &     1.6326 \times 10^{(-07)}  &     3.2653 \times 10^{(-07)} \\
         s=15   &  0.0000   &     1.6326 \times 10^{(-07)}  &     3.2653 \times 10^{(-07)}  \\
         s=16   &  0.0000   &       8.5067                  &     9.5047 \\
         s=17   &  0.0000   &       8.5067                  &     9.5047 \\
\end{block}
\end{blockarray}
\]

In Section~\ref{normal_comparison_small}, the transition probabilities in $P$ are assigned randomly. $P$ is given by
\[P=
\scriptsize
\begin{blockarray}{ccccccc}
            & w=0    &w=1     & w=2     &w=3     & w=4    &w=5   \\
\begin{block}{c(cccccc)}
    w=0     & 0.1697 & 0.1663 & 0.1662  & 0.1629 & 0.1691 & 0.1658 \\
    w=1     & 0.1584 & 0.1776 & 0.1552  & 0.1739 & 0.1579 & 0.1770 \\
    w=2     & 0.1689 & 0.1656 & 0.1651  & 0.1619 & 0.1709 & 0.1676 \\
    w=3     & 0.1577 & 0.1768 & 0.1542  & 0.1728 & 0.1596 & 0.1789 \\
    w=4     & 0.1679 & 0.1645 & 0.1635  & 0.1602 & 0.1737 & 0.1702 \\
    w=5     & 0.1567 & 0.1757 & 0.1526  & 0.1711 & 0.1621 & 0.1818 \\
\end{block}
\end{blockarray}
\]

For the case of rarely visited experiment in Section~\ref{comp_rar}, $P$ is given by
\[P=
\scriptsize
\begin{blockarray}{ccccccc}
            & w=0    &w=1     & w=2     &w=3     & w=4    &w=5   \\
\begin{block}{c(cccccc)}
    w=0     & 0.3690 & 0.0256 & 0.5228  & 0.0363 & 0.0434 & 0.0029 \\
    w=1     & 0.3834 & 0.0112 & 0.5432  & 0.0159 & 0.0451 & 0.0012 \\
    w=2     & 0.3876 & 0.0269 & 0.5114  & 0.0356 & 0.0360 & 0.0025 \\
    w=3     & 0.4027 & 0.0118 & 0.5314  & 0.0156 & 0.0374 & 0.0011 \\
    w=4     & 0.5165 & 0.0359 & 0.3401  & 0.0236 & 0.0784 & 0.0055 \\
    w=5     & 0.5367 & 0.0157 & 0.3533  & 0.0104 & 0.0814 & 0.0025 \\
\end{block}
\end{blockarray}
\]

\subsection{Large Scenario - 364 States}
The large scenario with 364 states was implemented as follows. The set of the states is
given by $S=\{0,...,363\}$. The states evolves according to:
{
\small
\begin{numcases}{s'=}
   52\floor*{\frac{s}{52}}-52a+4w+z, & $s$  is even or 0  \nonumber \\
   52\floor*{\frac{s}{52}}+4w+z, & $s$  is odd and greater than 312 and  $a=0$ \nonumber \\
   52\ceil*{\frac{s}{52}}-52a+4w+z, & else  \label{negative_364}
\end{numcases}
} where $w$ and $z$ are the disturbances at state $s$.
The disturbances are modeled by two independent Markov
processes with values $\mathcal{W}=\{0,...,12\}$ and $\mathcal{Z}=\{0,...,3\}$,
and transition probability matrices $P_w$ and $P_z$, respectively.

The transition probability from the current state $s$ to next state $s'$, given that action $a$
is selected, is given by
{
\small
\begin{equation} \label{state_transition_prob}
p(s'|s,a)= \begin{cases}
      p(w|w^p) * p(z|z^p), & \text{if}~ \eqref{negative_364}~ \text{is satisfied} \\
      0, & \text{else}
   \end{cases}
\end{equation}
}
where $w$, $z$, $w^p$, and $z^p$ are the current and the previous values of the two disturbances, respectively.

The set of actions at state $s$ is given by $\mathcal{A}^s=\{0,..,a^s_{\max}\}$,
where $a^s_{\max}=\floor*{\frac{s}{52}}$.

In Section~\ref{normal_comparison_small}, the transition probabilities in $P_w$ and $P_z$ are assigned randomly, and given by

\setcounter{MaxMatrixCols}{20}
\begin{equation} P_w=
\scriptsize
\begin{bmatrix}
    0.0882  &  0.0652  &  0.0572  &  0.1050  &  0.0849  &  0.0283  &  0.0439  &  0.1042  &  0.1078  &  0.0216  &  0.0392  &  0.1926  &  0.0619 \\
    0.1527  &  0.1975  &  0.0088  &  0.0125  &  0.0273  &  0.1800  &  0.0414  &  0.0287  &  0.1984  &  0.0320  &  0.0661  &  0.0040  &  0.0506 \\
    0.1158  &  0.0573  &  0.0571  &  0.1142  &  0.0931  &  0.1123  &  0.0867  &  0.0844  &  0.0075  &  0.0473  &  0.0755  &  0.0498  &  0.0990 \\
    0.0088  &  0.1145  &  0.1321  &  0.1760  &  0.1015  &  0.1220  &  0.0067  &  0.0527  &  0.0536  &  0.0589  &  0.0105  &  0.0975  &  0.0652 \\
    0.0857  &  0.0142  &  0.0767  &  0.0221  &  0.0915  &  0.1323  &  0.0728  &  0.0267  &  0.0338  &  0.1240  &  0.0902  &  0.1091  &  0.1209 \\
    0.0088  &  0.0895  &  0.0743  &  0.1318  &  0.0726  &  0.0359  &  0.0576  &  0.1266  &  0.0935  &  0.0723  &  0.0852  &  0.0452  &  0.1067 \\
    0.0744  &  0.0125  &  0.1310  &  0.1235  &  0.0719  &  0.0256  &  0.1480  &  0.0621  &  0.1316  &  0.1118  &  0.0991  &  0.0043  &  0.0042 \\
    0.1324  &  0.1089  &  0.1001  &  0.0452  &  0.0708  &  0.0666  &  0.0298  &  0.0070  &  0.0751  &  0.1295  &  0.1374  &  0.0672  &  0.0300 \\
    0.1755  &  0.0346  &  0.0330  &  0.1280  &  0.0981  &  0.1472  &  0.0065  &  0.1520  &  0.0026  &  0.0535  &  0.1174  &  0.0359  &  0.0157 \\
    0.0243  &  0.1317  &  0.0349  &  0.0562  &  0.0440  &  0.0712  &  0.1181  &  0.0754  &  0.1059  &  0.1294  &  0.0895  &  0.0945  &  0.0249 \\
    0.1081  &  0.0274  &  0.0277  &  0.1634  &  0.1636  &  0.1346  &  0.0425  &  0.0074  &  0.0322  &  0.0812  &  0.1047  &  0.1019  &  0.0053 \\
    0.0019  &  0.0473  &  0.1267  &  0.1174  &  0.0203  &  0.0445  &  0.0377  &  0.0965  &  0.1570  &  0.1442  &  0.1210  &  0.0335  &  0.0520 \\
    0.0628  &  0.0098  &  0.1174  &  0.1065  &  0.0664  &  0.0385  &  0.0296  &  0.0828  &  0.1785  &  0.0289  &  0.0379  &  0.1003  &  0.1406
\end{bmatrix}
\end{equation}

\[
\scriptsize
P_z=\begin{bmatrix}
    0.3338  &  0.2012  &  0.3305  &  0.1345 \\
    0.3671  &  0.1581  &  0.1364  &  0.3384 \\
    0.1944  &  0.4143  &  0.2839  &  0.1074 \\
    0.0063  &  0.3457  &  0.2206  &  0.4274
\end{bmatrix}
\]

For the case of rarely visited experiment in Section~\ref{comp_rar}, $P_w$ and $P_z$ are given by

\setcounter{MaxMatrixCols}{20}
\begin{equation}
\scriptsize
P_w=\begin{bmatrix}
    0.1901 &   0.1054  &  0.1886  &  0.1543 &   0.0946  &  0.0023  &  0.0041  &  0.2048  &  0.0003  &  0.0042  &  0.0050  &  0.0200  &  0.0263 \\
    0.2400 &   0.1579  &  0.1899  &  0.0861 &   0.0421  &  0.0054  &  0.0046  &  0.1776  &  0.0062  &  0.0062  &  0.0001  &  0.0400  &  0.0439 \\
    0.2101 &   0.0998  &  0.1952  &  0.2003 &   0.1002  &  0.0031  &  0.0012  &  0.1467  &  0.0022  &  0.0022  &  0.0005  &  0.0200  &  0.0185 \\
    0.2409 &   0.1292  &  0.1930  &  0.0861 &   0.0708  &  0.0036  &  0.0140  &  0.1666  &  0.0060  &  0.0055  &  0.0004  &  0.0500  &  0.0339 \\
    0.2003 &   0.1749  &  0.1889  &  0.1563 &   0.0251  &  0.0054  &  0.0055  &  0.1907  &  0.0065  &  0.0077  &  0.0002  &  0.0211  &  0.0174 \\
    0.1809 &   0.1054  &  0.1974  &  0.1531 &   0.0950  &  0.0023  &  0.0012  &  0.2053  &  0.0090  &  0.0047  &  0.0003  &  0.0221  &  0.0233 \\
    0.2401 &   0.1567  &  0.1901  &  0.0852 &   0.0471  &  0.0044  &  0.0056  &  0.1720  &  0.0081  &  0.0035  &  0.0050  &  0.0420  &  0.0402 \\
    0.2037 &   0.1709  &  0.1881  &  0.1579 &   0.0248  &  0.0061  &  0.0068  &  0.1917  &  0.0043  &  0.0064  &  0.0015  &  0.0199  &  0.0179 \\
    0.1888 &   0.1056  &  0.1879  &  0.1539 &   0.0955  &  0.0043  &  0.0039  &  0.2033  &  0.0017  &  0.0053  &  0.0052  &  0.0189  &  0.0257 \\
    0.2422 &   0.1303  &  0.1941  &  0.0859 &   0.0742  &  0.0047  &  0.0033  &  0.1699  &  0.0069  &  0.0049  &  0.0016  &  0.0421  &  0.0399 \\
    0.1895 &   0.1063  &  0.1879  &  0.1553 &   0.0939  &  0.0024  &  0.0039  &  0.2033  &  0.0006  &  0.0050  &  0.0061  &  0.0199  &  0.0259 \\
    0.2394 &   0.0901  &  0.1888  &  0.0740 &   0.1089  &  0.0059  &  0.0050  &  0.1912  &  0.0081  &  0.0049  &  0.0005  &  0.0405  &  0.0427 \\
    0.2000 &   0.1198  &  0.1800  &  0.1004 &   0.0802  &  0.0082  &  0.0062  &  0.1633  &  0.0473  &  0.0051  &  0.0056  &  0.0400  &  0.0439
\end{bmatrix}
\end{equation}

\[
\scriptsize
P_z=\begin{bmatrix}
    0.9000  &  0.0650  &  0.0300  &  0.0050 \\
    0.9215  &  0.0285  &  0.0250  &  0.0250 \\
    0.9000  &  0.0650  &  0.0300  &  0.0050 \\
    0.9215  &  0.0285  &  0.0250  &  0.0250
\end{bmatrix}
\]

} 

The reward matrix is a $364 \times 7$ matrix, where $364$ is the number of states and $7$ is the maximum
number of actions available at each state. The reward matrix used in Section~\ref{normal_comparison_small}
is expressed using one small matrix $X$ and Table~\ref{Table_reward_1}.
\setcounter{MaxMatrixCols}{52}
\begin{equation} \nonumber
\scriptsize
X=\begin{bmatrix}
    0.0000 &   0.0000  &  0.0000  &  0.0000 &   0.0000  &  0.0000  &  0.0000   \\
    0.0000 &   0.0000  &  0.0000  &  0.0000 &   0.0000  &  0.0000  &  0.0000   \\
    0.0000 &   0.0000  &  0.0000  &  0.0000 &   0.0000  &  0.0000  &  0.0000   \\
    0.0000 &   0.0000  &  0.0000  &  0.0000 &   0.0000  &  0.0000  &  0.0000   \\
    0.0000 &   0.0000  &  0.0000  &  0.0000 &   0.0000  &  0.0000  &  0.0000   \\
    0.0000 &   0.0000  &  0.0000  &  0.0000 &   0.0000  &  0.0000  &  0.0000   \\
    0.0000 &   0.0000  &  0.0000  &  0.0000 &   0.0000  &  0.0000  &  0.0000   \\
    0.0000 &   0.0000  &  0.0000  &  0.0000 &   0.0000  &  0.0000  &  0.0000   \\
    0.0000 &   0.0000  &  0.0000  &  0.0000 &   0.0000  &  0.0000  &  0.0000   \\
    0.0000 &   0.0000  &  0.0000  &  0.0000 &   0.0000  &  0.0000  &  0.0000   \\
    0.0000 &   0.0000  &  0.0000  &  0.0000 &   0.0000  &  0.0000  &  0.0000   \\
    0.0000 &   0.0000  &  0.0000  &  0.0000 &   0.0000  &  0.0000  &  0.0000   \\
    0.0000 &   0.0000  &  0.0000  &  0.0000 &   0.0000  &  0.0000  &  0.0000   \\
    0.0000 &   0.0000  &  0.0000  &  0.0000 &   0.0000  &  0.0000  &  0.0000   \\
    0.0000 &   0.0000  &  0.0000  &  0.0000 &   0.0000  &  0.0000  &  0.0000   \\
    0.0000 &   0.0000  &  0.0000  &  0.0000 &   0.0000  &  0.0000  &  0.0000   \\
    0.0000 &   0.0000  &  0.0000  &  0.0000 &   0.0000  &  0.0000  &  0.0000   \\
    0.0000 &   0.0000  &  0.0000  &  0.0000 &   0.0000  &  0.0000  &  0.0000   \\
    0.0000 &   0.0000  &  0.0000  &  0.0000 &   0.0000  &  0.0000  &  0.0000   \\
    0.0000 &   0.0000  &  0.0000  &  0.0000 &   0.0000  &  0.0000  &  0.0000   \\
    0.0000 &   0.0000  &  0.0000  &  0.0000 &   0.0000  &  0.0000  &  0.0000   \\
    0.0000 &   0.0000  &  0.0000  &  0.0000 &   0.0000  &  0.0000  &  0.0000   \\
    0.0000 &   0.0000  &  0.0000  &  0.0000 &   0.0000  &  0.0000  &  0.0000   \\
    0.0000 &   0.0000  &  0.0000  &  0.0000 &   0.0000  &  0.0000  &  0.0000   \\
    0.0000 &   0.0101  &  0.0201  &  0.0300 &   0.0398  &  0.0496  &  0.0593   \\
    0.0000 &   0.0101  &  0.0201  &  0.0300 &   0.0398  &  0.0496  &  0.0593   \\
    0.0000 &   0.0101  &  0.0201  &  0.0300 &   0.0398  &  0.0496  &  0.0593   \\
    0.0000 &   0.0101  &  0.0201  &  0.0300 &   0.0398  &  0.0496  &  0.0593   \\
    0.0000 &   0.0267  &  0.0530  &  0.0788 &   0.1041  &  0.1290  &  0.1535   \\
    0.0000 &   0.0267  &  0.0530  &  0.0788 &   0.1041  &  0.1290  &  0.1535  \\
    0.0000 &   0.0267  &  0.0530  &  0.0788 &   0.1041  &  0.1290  &  0.1535   \\
    0.0000 &   0.0267  &  0.0530  &  0.0788 &   0.1041  &  0.1290  &  0.1535   \\
    0.0000 &   0.0514  &  0.1010  &  0.1490 &   0.1955  &  0.2404  &  0.2841   \\
    0.0000 &   0.0514  &  0.1010  &  0.1490 &   0.1955  &  0.2404  &  0.2841   \\
    0.0000 &   0.0514  &  0.1010  &  0.1490 &   0.1955  &  0.2404  &  0.2841  \\
    0.0000 &   0.0514  &  0.1010  &  0.1490 &   0.1955  &  0.2404  &  0.2841  \\
    0.0000 &   4.0323  &  4.9875  &  5.5573 &   5.9646  &  6.2819  &  6.5419   \\
    0.0000 &   4.0323  &  4.9875  &  5.5573 &   5.9646  &  6.2819  &  6.5419   \\
    0.0000 &   4.0323  &  4.9875  &  5.5573 &   5.9646  &  6.2819  &  6.5419   \\
    0.0000 &   4.0323  &  4.9875  &  5.5573 &   5.9646  &  6.2819  &  6.5419   \\
    0.0000 &   6.1492  &  7.1390  &  7.7205 &   8.1339  &  8.4548  &  8.7171   \\
    0.0000 &   6.1492  &  7.1390  &  7.7205 &   8.1339  &  8.4548  &  8.7171   \\
    0.0000 &   6.1492  &  7.1390  &  7.7205 &   8.1339  &  8.4548  &  8.7171  \\
    0.0000 &   6.1492  &  7.1390  &  7.7205 &   8.1339  &  8.4548  &  8.7171   \\
    0.0000 &   9.0228  &  10.0214  &  10.6059 &   11.0207  &  11.3425  &  11.6055   \\
    0.0000 &   9.0228  &  10.0214  &  10.6059 &   11.0207  &  11.3425  &  11.6055   \\
    0.0000 &   9.0228  &  10.0214  &  10.6059 &   11.0207  &  11.3425  &  11.6055  \\
    0.0000 &   9.0228  &  10.0214  &  10.6059 &   11.0207  &  11.3425  &  11.6055  \\
    0.0000 &   9.6817  &  10.6808  &  11.2655 &   11.6804  &  12.0022  &  12.2652  \\
    0.0000 &   9.6817  &  10.6808  &  11.2655 &   11.6804  &  12.0022  &  12.2652   \\
    0.0000 &   9.6817  &  10.6808  &  11.2655 &   11.6804  &  12.0022  &  12.2652   \\
    0.0000 &   9.6817  &  10.6808  &  11.2655 &   11.6804  &  12.0022  &  12.2652
\end{bmatrix}
\end{equation}

\begin{table}[!h]
  \centering
  \scriptsize
  \begin{tabular}{|c|c|c|c|c|c|c|c|}
   \hline
   Actions              & $r(s,a=0)$  &  $r(s,a=1)$  &  $r(s,a=2)$  &  $r(s,a=3)$  &  $r(s,a=4)$  &  $r(s,a=5)$  &  $r(s,a=6)$  \\ \hline
   $s=\{0,...,51\}$     &   $X(:,1)$  &      -       &      -       &      -       &      -       &      -       &      -      \\ \hline
   $s=\{52,...,103\}$   &   $X(:,1)$  &   $X(:,2)$   &      -       &      -       &      -       &      -       &      -      \\ \hline
   $s=\{104,...,155\}$  &   $X(:,1)$  &   $X(:,2)$   &   $X(:,3)$   &      -       &      -       &      -       &      -      \\ \hline
   $s=\{156,...,207\}$  &   $X(:,1)$  &   $X(:,2)$   &   $X(:,3)$   &   $X(:,4)$   &      -       &      -       &      -      \\ \hline
   $s=\{208,...,259\}$  &   $X(:,1)$  &   $X(:,2)$   &   $X(:,3)$   &   $X(:,4)$   &   $X(:,5)$   &      -       &      -      \\ \hline
   $s=\{260,...,311\}$  &   $X(:,1)$  &   $X(:,2)$   &   $X(:,3)$   &   $X(:,4)$   &   $X(:,5)$   &   $X(:,6)$   &      -      \\ \hline
   $s=\{312,...,363\}$  &   $X(:,1)$  &   $X(:,2)$   &   $X(:,3)$   &   $X(:,4)$   &   $X(:,5)$   &   $X(:,6)$   &   $X(:,7)$  \\ \hline
   \end{tabular}
   \caption{The immediate rewards for all state-action pairs.} \label{Table_reward_1}
\end{table}

\newpage

The reward matrix used in Section~\ref{comp_rar} is expressed using
one small matrix $Y$ and Table~\ref{Table_reward_2}.

\setcounter{MaxMatrixCols}{52}
\begin{equation} \nonumber
\scriptsize
Y=\begin{bmatrix}
    0.0000 &   0.0000  &  0.0000  &  0.0000 &   0.0000  &  0.0000  &  0.0000   \\
    0.0000 &   0.0000  &  0.0000  &  0.0000 &   0.0000  &  0.0000  &  0.0000   \\
    0.0000 &   0.0000  &  0.0000  &  0.0000 &   0.0000  &  0.0000  &  0.0000   \\
    0.0000 &   0.0000  &  0.0000  &  0.0000 &   0.0000  &  0.0000  &  0.0000   \\
    0.0000 &   0.0000  &  0.0000  &  0.0000 &   0.0000  &  0.0000  &  0.0000   \\
    0.0000 &   0.0000  &  0.0000  &  0.0000 &   0.0000  &  0.0000  &  0.0000   \\
    0.0000 &   0.0000  &  0.0000  &  0.0000 &   0.0000  &  0.0000  &  0.0000   \\
    0.0000 &   0.0000  &  0.0000  &  0.0000 &   0.0000  &  0.0000  &  0.0000   \\
    0.0000 &   0.0000  &  0.0000  &  0.0000 &   0.0000  &  0.0000  &  0.0000   \\
    0.0000 &   0.0000  &  0.0000  &  0.0000 &   0.0000  &  0.0000  &  0.0000   \\
    0.0000 &   0.0000  &  0.0000  &  0.0000 &   0.0000  &  0.0000  &  0.0000   \\
    0.0000 &   0.0000  &  0.0000  &  0.0000 &   0.0000  &  0.0000  &  0.0000   \\
    0.0000 &   0.0000  &  0.0000  &  0.0000 &   0.0000  &  0.0000  &  0.0000   \\
    0.0000 &   0.0000  &  0.0000  &  0.0000 &   0.0000  &  0.0000  &  0.0000   \\
    0.0000 &   0.0000  &  0.0000  &  0.0000 &   0.0000  &  0.0000  &  0.0000   \\
    0.0000 &   0.0000  &  0.0000  &  0.0000 &   0.0000  &  0.0000  &  0.0000   \\
    0.0000 &   0.0000  &  0.0000  &  0.0000 &   0.0000  &  0.0000  &  0.0000   \\
    0.0000 &   0.0000  &  0.0000  &  0.0000 &   0.0000  &  0.0000  &  0.0000   \\
    0.0000 &   0.0000  &  0.0000  &  0.0000 &   0.0000  &  0.0000  &  0.0000   \\
    0.0000 &   0.0000  &  0.0000  &  0.0000 &   0.0000  &  0.0000  &  0.0000   \\
    0.0000 &   0.0000  &  0.0000  &  0.0000 &   0.0000  &  0.0000  &  0.0000   \\
    0.0000 &   0.0000  &  0.0000  &  0.0000 &   0.0000  &  0.0000  &  0.0000   \\
    0.0000 &   0.0000  &  0.0000  &  0.0000 &   0.0000  &  0.0000  &  0.0000   \\
    0.0000 &   0.0000  &  0.0000  &  0.0000 &   0.0000  &  0.0000  &  0.0000   \\
    0.0000 &   0.0000  &  0.0000  &  0.0000 &   0.0000  &  0.0000  &  0.0000   \\
    0.0000 &   0.0000  &  0.0000  &  0.0000 &   0.0000  &  0.0000  &  0.0000   \\
    0.0000 &   0.0000  &  0.0000  &  0.0000 &   0.0000  &  0.0000  &  0.0000   \\
    0.0000 &   0.0000  &  0.0000  &  0.0000 &   0.0000  &  0.0000  &  0.0000   \\
    0.0000 &   0.0000  &  0.0000  &  0.0000 &   0.0000  &  0.0000  &  0.0000   \\
    0.0000 &   0.0000  &  0.0000  &  0.0000 &   0.0000  &  0.0000  &  0.0000  \\
    0.0000 &   0.0000  &  0.0000  &  0.0000 &   0.0000  &  0.0000  &  0.0000   \\
    0.0000 &   0.0000  &  0.0000  &  0.0000 &   0.0000  &  0.0000  &  0.0000   \\
    0.0000 &   0.7653  &  1.2627  &  1.6319 &   1.9256  &  2.1695  &  2.3781   \\
    0.0000 &   0.7653  &  1.2627  &  1.6319 &   1.9256  &  2.1695  &  2.3781   \\
    0.0000 &   0.7653  &  1.2627  &  1.6319 &   1.9256  &  2.1695  &  2.3781  \\
    0.0000 &   0.7653  &  1.2627  &  1.6319 &   1.9256  &  2.1695  &  2.3781  \\
    0.0000 &   10.5861  &  11.5857  &  12.1705 &   12.5854  &  12.9073  &  13.1703   \\
    0.0000 &   10.5861  &  11.5857  &  12.1705 &   12.5854  &  12.9073  &  13.1703   \\
    0.0000 &   10.5861  &  11.5857  &  12.1705 &   12.5854  &  12.9073  &  13.1703   \\
    0.0000 &   10.5861  &  11.5857  &  12.1705 &   12.5854  &  12.9073  &  13.1703   \\
    0.0000 &   19.4164  &  20.4164  &  21.0014 &   21.4164  &  21.7384  &  22.0014   \\
    0.0000 &   19.4164  &  20.4164  &  21.0014 &   21.4164  &  21.7384  &  22.0014   \\
    0.0000 &   19.4164  &  20.4164  &  21.0014 &   21.4164  &  21.7384  &  22.0014  \\
    0.0000 &   19.4164  &  20.4164  &  21.0014 &   21.4164  &  21.7384  &  22.0014   \\
    0.0000 &   21.7904  &  22.7904  &  23.3754 &   23.7904  &  24.1123  &  24.3754   \\
    0.0000 &   21.7904  &  22.7904  &  23.3754 &   23.7904  &  24.1123  &  24.3754   \\
    0.0000 &   21.7904  &  22.7904  &  23.3754 &   23.7904  &  24.1123  &  24.3754  \\
    0.0000 &   21.7904  &  22.7904  &  23.3754 &   23.7904  &  24.1123  &  24.3754  \\
    0.0000 &   22.9676  &  23.9676  &  24.5526 &   24.9676  &  25.2896  &  25.5526  \\
    0.0000 &   22.9676  &  23.9676  &  24.5526 &   24.9676  &  25.2896  &  25.5526   \\
    0.0000 &   22.9676  &  23.9676  &  24.5526 &   24.9676  &  25.2896  &  25.5526   \\
    0.0000 &   22.9676  &  23.9676  &  24.5526 &   24.9676  &  25.2896  &  25.5526
\end{bmatrix}
\end{equation}

\begin{table}[!h]
  \centering
  \scriptsize
  \begin{tabular}{|c|c|c|c|c|c|c|c|}
   \hline
   Actions              & $r(s,a=0)$  &  $r(s,a=1)$  &  $r(s,a=2)$  &  $r(s,a=3)$  &  $r(s,a=4)$  &  $r(s,a=5)$  &  $r(s,a=6)$  \\ \hline
   $s=\{0,...,51\}$     &   $Y(:,1)$  &      -       &      -       &      -       &      -       &      -       &      -      \\ \hline
   $s=\{52,...,103\}$   &   $Y(:,1)$  &   $Y(:,2)$   &      -       &      -       &      -       &      -       &      -      \\ \hline
   $s=\{104,...,155\}$  &   $Y(:,1)$  &   $Y(:,2)$   &   $Y(:,3)$   &      -       &      -       &      -       &      -      \\ \hline
   $s=\{156,...,207\}$  &   $Y(:,1)$  &   $Y(:,2)$   &   $Y(:,3)$   &   $Y(:,4)$   &      -       &      -       &      -      \\ \hline
   $s=\{208,...,259\}$  &   $Y(:,1)$  &   $Y(:,2)$   &   $Y(:,3)$   &   $Y(:,4)$   &   $Y(:,5)$   &      -       &      -      \\ \hline
   $s=\{260,...,311\}$  &   $Y(:,1)$  &   $Y(:,2)$   &   $Y(:,3)$   &   $Y(:,4)$   &   $Y(:,5)$   &   $Y(:,6)$   &      -      \\ \hline
   $s=\{312,...,363\}$  &   $Y(:,1)$  &   $Y(:,2)$   &   $Y(:,3)$   &   $Y(:,4)$   &   $Y(:,5)$   &   $Y(:,6)$   &   $Y(:,7)$  \\ \hline
   \end{tabular}
   \caption{The immediate rewards for all state-action pairs.} \label{Table_reward_2}
\end{table}

\end{document}